# SLJP: Semantic Extraction based Legal Judgment Prediction

Prameela Madambakam[1], Shathanaa Rajmohan[2], Himangshu Sharma[1], Tummepalli Anka Chandrahas Purushotham Gupta[1]

[1]Department of Computer Science and Engineering, Indian Institute of Information Technology, Sricity, India

[2]Dept of CSE, College of Engineering Guindy, Anna University, Chennai

*Abstract*— Legal Judgment Prediction (LJP) is a judicial assistance system that recommends the legal components such as applicable statues, prison term and penalty term by analyzing the given input case document. Indian legal system is in the need of technical assistance such as artificial intelligence to solve the crores of pending cases in various courts for years and its being increased day to day. Most of the existing Indian models did not adequately concentrate on the semantics embedded in the fact description (FD) that impacts the decision. The proposed semantic extraction based LJP (SLJP) model provides the advantages of pretrained transformers for complex unstructured legal case document understanding and to generate embeddings. The model draws the in-depth semantics of the given FD at multiple levels i.e., chunk and case document level by following the divide and conquer approach. It creates the concise view of the given fact description using the extracted semantics as per the original court case document structure and predicts judgment using attention mechanism. We tested the model performance on two available Indian datasets Indian Legal Documents corpus (ILDC) and Indian Legal Statue Identification (ILSI) and got promising results. Also shown the highest performance and less performance degradation for increased epochs than base models on ILDC dataset.

*Keywords*— concise semantic extraction, legal judgment prediction, criminal cases, divide and conquer, multi-label classification, binary classification, statue identification

## I. Introduction

The main role of LJP is to predict legal decisions as delivered by the judges in court by following the Indian Penal Code (IPC) sections after analyzing the input case document referred as fact description. It acts as a reference system to the judges in delivering the judgment, lawyers to argue the case and end users to know the possible prison and penalty terms [1].

Here we are considering only criminal cases which can be solved by applying the IPC sections whereas solving of the civil cases systematically is difficult because of its heterogeneous nature. Generally legal judgement consists of the multiple components named as subtasks, e.g., law sections, prison term, and penalty term prediction. Based on the dataset availability, SLJP solved only law article prediction by applying the multi-label classification techniques. It plays a vital role in delivering the judgment and it intern determines the possible prison and penalty term imposed on the criminal based on the requirements. In real time, judge follows a procedure in deciding the verdict. He first determines the applicable IPC sections after thorough understanding of the given case document. He did this by analyzing the semantics embedded in the FD illustrated by the sequence of events happened and its intensity indicated by the loss and damages occurred. Later he refers the IPC to find the applicability and limits of the prison and penalty term for the corresponding law section and make a final decision based on the semantics analyzed and evidences proved. E.g., 1. Section 379 in The Indian Penal Code describes, "Punishment for theft — Whoever commits theft shall be punished with imprisonment of either description for a term which may extend to three years, or with fine, or with both." 2. Section 363 in The Indian Penal Code describes, "Punishment for kidnapping— Whoever kidnaps any person from [India] or from lawful guardianship, shall be punished with imprisonment of either description for a term which may extend to seven years, and shall also be liable to fine" [1].

Hence determining the judgment for the given case document mainly depends on the semantics involved in the given FD explained by the various crime events involved and evidences proved. These semantics is mapped with the IPC section descriptions to identify the applicable statues. So, extraction of these semantics from the given fact description plays a significant part to judge the case with accurate decision. Most of the existing methodologies especially Indian legal works did not adequately concentrate on this semantics extraction during judgment decision process which is addressed here. This work proposed a novel semantics extraction based legal judgment prediction method, to extract in-depth semantics of the given case FD at multiple levels i.e., chunk and case document level by following the divide and conquer approach. Later it creates a concise view of the fact description using these extracted semantics as per the original court case document structure having document summary at the last few lines. Then it predicts the decision using attention mechanism.

Semantics based prediction approach followed here helps in understanding the given case document through transformer ex: XLNET and draw chunk level semantics by applying the divide and conquer approach and extract document level semantics form aggregated chunk semantics dynamically. To avoid the data loss occurred because of extracting semantics at multiple levels it introduced the Concise Extraction (CE) module. This module enhances the semantics from both aggregated chunk and document semantics by creating the concise view of the FD with these extracted semantics and apply the attention mechanism. The resultant enriched semantics extracted by following the proposed framework contains the essential key phrases that assists the subsequent prediction module in making the final decision. One of the drawbacks for Indian LJP is to make analysis of the complex unstructured case documents which are not in a similar format. This formatting is a consequence of Indian judges do not follow any particular format in producing the judgment copy. The advantage of SLJP model is its ability to understand these complex unstructured documents through transformer and create a concise view of the case document before predicting the judgment. SLJP is evaluated on several legal baseline models and got promising results.

---

[1]https://www.indiacode.nic.in/bitstream/123456789/15289/1/ipc_act.pdf

The structure of this article is formatted as follows: section 2 explains the related work, section 3 summarizes the problem statement, section 4 illustrates the proposed work, section 5 represents the results and experimental analysis, section 6 describes the ablation analysis, and lastly section 7 discusses the conclusion and future work.

## II. RELATED WORK

*Multi-tasking:*

Yao et al. [2] worked on the multi-tasking model and proposed a GHEDAP method to automatically identify the various subtasks involved in the FD and dependencies between them dynamically. It also extracts the in-depth semantics embedded in the FD. Yao et al. [3] proposed CSDNET as an extension of GHEDAP by identifying the commonalities, specificities and dependencies among the various subtasks involved in the FD with the support of three modules named learning, denoising and reinforcing module.

*Datasets:*

Paul et al. [4] proposed LeSICiN, a heterogeneous graph model that explores both text and the legal citation network for the legal statue identification task. He created a dataset named ILSI for this task, by extracting case documents from several Indian courts and sections from IPC. Malik et al. [5] created a legal corpus named ILDC with 35k cases from supreme court of India mapped with original judgments. He then introduced CJPE a hierarchical model for judgment prediction and to deliver explanation.

*Embeddings: Encoding methods*

Pagliardini et al. [6] proposed an unsupervised model sent2vec inspired by C-BOW to produce sentence embeddings using word vectors in addition to n-grams. It efficiently learns from large datasets in a streaming fashion and infer embeddings for new sentences. Le et al. [7] derived an unsupervised technique called paragraph vector, which output fixed-length features for variable-length inputs of paragraphs, documents, and sentences. Trained document dense vector representation is used to predict words of the document. Pennington et al. [8] explore the benefits of global matrix factorization and local context window methods to produce an unsupervised log-bilinear regression model that results in meaningful substructure vector space.

*Embeddings: Transformer models*

Devlin et al. [9] adapted the masked language models and next sentence prediction task to build pretrained deep bidirectional representation model BERT as the first finetuning model with just single output layer added. Yang et al. [10] combined Transformer-XL and the autoregressive model into a pretraining model XLNet. It avoids the limitations of BERT named pretrain-finetune discrepancy through the segment recurrence mechanism and relative encoding scheme.

*Baselines of ILSI dataset:*

Luo et al. [11] modeled an unified framework for charge prediction and the relevant article extraction task using a two-stack attention-based neural network implemented through document and article encoder. Wang et al. [12] worked on DPAM to tackle the two aspects of crime classification listed as label dynamic problem and the label imbalance problem. It improves the performance by exploring the information returned by the multi-task learning and the dynamic threshold predictor. Wang et al. [13] fused the hierarchical structure and label semantics into a unified framework named HMN. It converts classification task into a matching problem by decomposing article definitions into the residual and alignment sub structures for good prediction results. Xu et al. [14] developed LADAN to address the problem of confusing law articles in LJP through graph distillation operator that effectively mine the distinguishing features of the confusing law articles. Chalkidis et al. [15] contributed a new english LJP dataset with the European Court of Human Rights (ECHR) cases. He derived a hierarchical version of BERT that overcomes the BERT length constraint and gives the best performance. Chalkidis et al. [16] adapted BERT in legal domain through systematic investigation of original BERT out of the box, further pre-train, and pre-train BERT from scratch on customized domains. He released LEGAL-BERT, LEGAL-BERT-SC, and LEGAL-BERT-SMALL which achieved more efficient and comparable performance than larger models. Hao et al. [17] designed DEAL for inductive link prediction between the nodes having only attribute information. He adapted the three components containing two node embedding encoders and one alignment mechanism.

## III. PROBLEM STATEMENT

The main task of SLJP is to predict judicial judgment by extracting semantics at multiple levels from the given FD.

*Multi-label classification:*

Generally, LJP comprises of several subtasks e.g., law articles, prison term and penalty term. As there is no multi-tasking dataset available for Indian context to predict three terms, this work limited to available ILSI dataset for identifying applicable law statues of the given criminal case document. It is a multi-label classification where each case document is mapped with multiple sections simultaneously which are applicable based on the multiple events involved. In real case, final prison and penalty term is the sum of prison and penalty terms of all events explained by multiple sections predicted or identified.

ILSI dataset contains cases belongs to top 100 frequently occurred sections or labels indicated by $L = \{k_1, k_2, k_3 \ldots k_{100}\}$. Given the FD $X$, SLJP model $S$ is trained to identify the corresponding labels i.e., applicable law sections for $X$ indicated by $Y_x$:

$$S(X) \rightarrow Y_x \quad (1)$$

here $Y_x = \{0,1\}^{|L|}$ indicates the set of $|L| = 100$ predicted labels corresponding to 100 sections in the dataset. $Y_x[k] \in \{0,1\}$ marks the model prediction of applicability of label $k$ to the case $X$. If the FD $X$ contains $n$ words, then $X = \{x_i\}_{i=1}^n$ then SLJP is formulized as:

$$S(\{x_i\}_{i=1}^n) = \{0,1\}^{|L|} \quad (2)$$

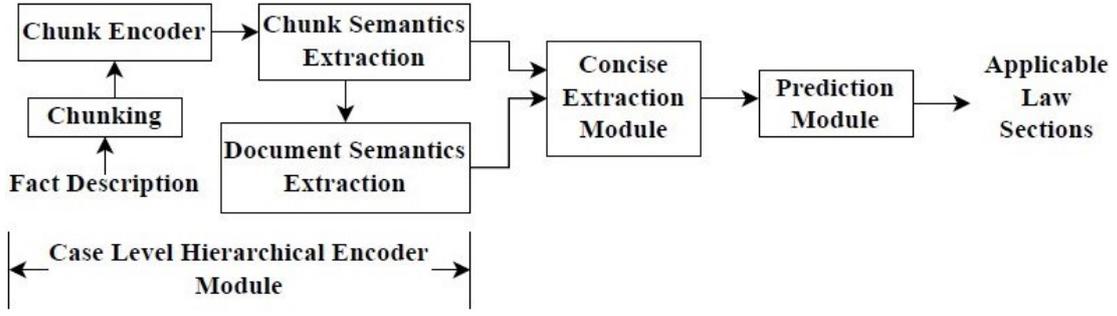

**Figure 1:** Overview of the Proposed Architecture

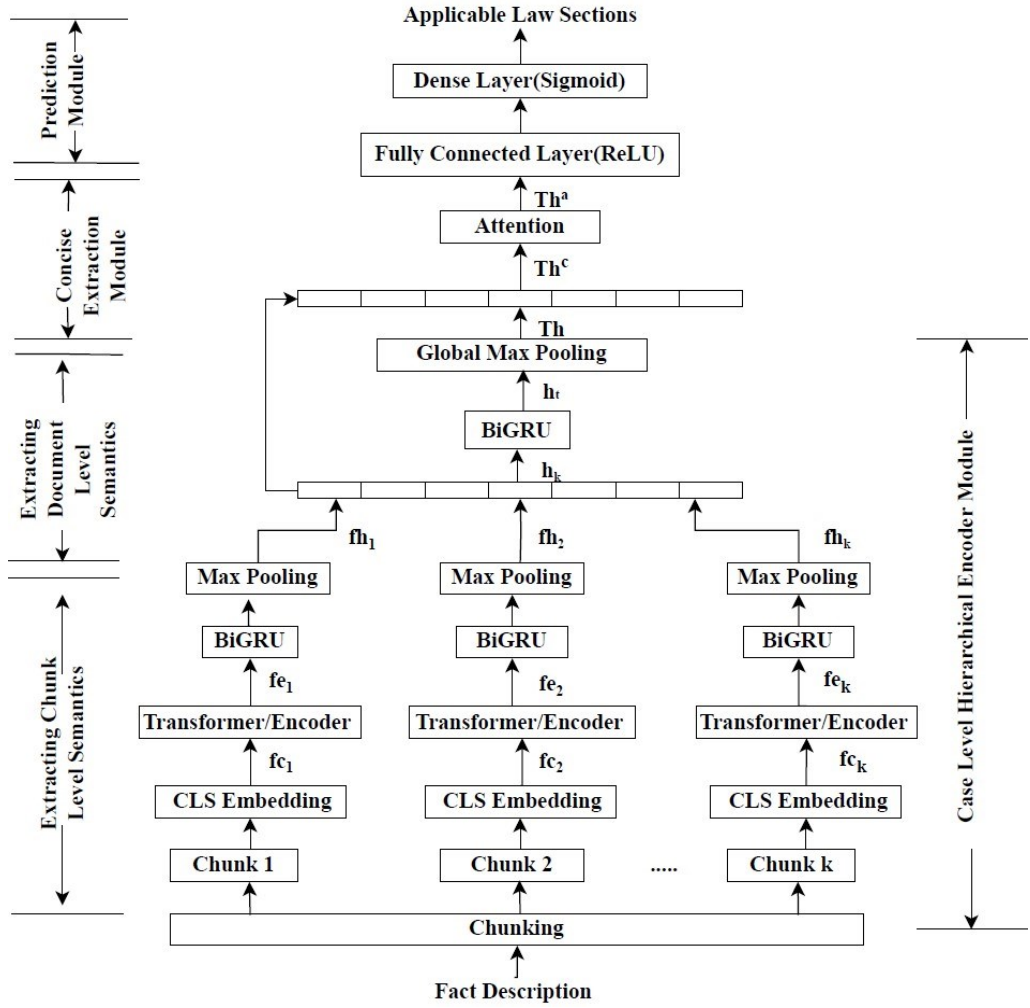

**Figure 2:** Detailed Architecture of the Proposed Framework

*Binary classification:*

Also applied SLJP model to predict the judgment on the binary classification dataset of ILDC supreme court case proceedings where outcome is either accept or reject. Given the fact description $X$, SLJP model $S$ is trained to predict the corresponding label i.e, acceptance status of $X$ indicated by $Y_x$:

$$S(X) \to Y_x \quad (3)$$

Here $Y_x = \{0,1\}$ indicates the predicted label of acceptance. If the FD contains $n$ words, then $X = \{x_i\}_{i=1}^{n}$, and SLJP is formulated as:

$$S(\{x_i\}_{i=1}^{n}) = \{0,1\} \quad (4)$$

## IV. PROPOSED WORK

Most of the Indian works are concentrating on the pretrained language models for case document understanding based judgment prediction. Chinese models are following the trend of semantic representation for improved performance in decision prediction. The proposed semantics based legal judgment prediction model combines the advantages of these 2 approaches: 1. uses pretrained language called transformers for legal case document understanding and generate embeddings and 2. draws the in-depth semantics of the given FD at multiple levels i.e., chunk and document level by following the divide and conquer approach. It then creates a concise view of the FD using these extracted semantics as per the original court case document structure having document summary at the last few lines and predicts judgment using attention mechanism.

Figure 1 shows the brief overview of the proposed architecture. It consists of three modules listed as case level hierarchical encoder, concise extraction module, and prediction module. First module case level hierarchical encoder dynamically extracts semantics at multiple levels i.e, chunk and document level by following the divide and conquer approach. It is designed to return the two outcomes and have multiple components build in it. The first component *chunking* takes the input fact description and iteratively segments it into the fixed sized chunks and then add CLS embedding for each chunk as delimiter. The resultant fixed sized chunks are then passed to the *Chunk Encoder* component which might be a transformer such as XLNet or encoder and return the embeddings for the input chunks. The resultant chunk embeddings are then passed to the *Chunk Semantics Extraction* component to get individual chunk semantics which is the first outcome of the case level hierarchical encoder. These individual chunk embeddings are concatenated to form aggregated chunk semantics and passed to the *Document Semantics Extraction* component. This component extracts document level in-depth semantics from aggregated chunk semantics which is second outcome of this module.

To avoid the data loss occurred because of extracting semantics at multiple levels SLJP introduced a novel second module named *Concise Extraction* (CE). It enhances the semantics from both aggregated chunk and document semantics by creating the concise view of the fact description as per the original court case document structure. It appends the document summary at the last few lines and applies attention to concentrate on the main words impacting subsequent prediction module. The last module named *Prediction* uses the resultant enriched semantics containing the essential key phrases impacting decision making by predicting the probabilities of the labels.

One of the drawbacks for Indian LJP is to make analysis of the complex unstructured case documents which are not in a similar format. This formatting is a consequence of Indian judges do not follow any particular format in producing the judgment copy. The advantage of SLJP model is its ability to understand these complex unstructured documents and predict the judgment. Figure 2 represents the detailed architecture of the proposed frame work.

### 4.1 Case level hierarchical encoder architecture

**Table 1:** Algorithm of case level hierarchical encoder module

---

**Algorithm1:** Case level hierarchical encoder

**Input:** Fact description

**Output:** Extracted case level in-depth semantic information at multiple levels such as chunk and document level towards verdict prediction task.

**Procedure:**

1. Input fact description $X = \{x_i\}_{i=1}^n$, where $n$ = number of words in FD.

2. **Chunking:** Divides $n$ length FD into $K$ chunks $fC_1, fC_2, \ldots, fC_k$ where each chunk size is $m \leq 512$ words for transformers and varies for encoders. Later adds CLS embedding as chunk delimiter.

3. **Chunk Processing:** For each chunk $fC_i$, where $i = 1, 2, \ldots k$ do the following.

   a. *Chunk encoding:* Input chunk $fC_i$, to pretrained transformer model such as XLNET or encoding method to get corresponding word embeddings denoted by $fe_i$.

   b. *Check semantics extraction:*

   i. Process each word embedding $x_{i,t}$ in chunk $fe_i$ with BiGRU to get $h_{i,t}$ containing comprehensive semantics of a specific chunk.

   $$h_{i,t} = BiGRU(x_{i,t}, h_{i,t-1}) \quad (5)$$

   ii. Concise chunk specific semantics in $h_{i,t}$ by applying Maxpooling to make it more informative and forms the individual chunk embeddings which is the first outcome of this module.

   $$fh_i = MaxPooling(h_{i,t}) \quad (6)$$

   c. Combine all individual chunk semantics $fh_1, fh_2, \ldots, fh_k$ to obtain aggregated chunk level semantics.

   $$h_k = \{fh_i\}_{i=1}^k \quad (7)$$

4. **Document semantics extraction:**

   i. Process each vector $x_t$ in $h_k$ with BiGRU to produce aggregated document or case level task specific semantics denoted by $h_t$.

   $$h_t = BiGRU(x_t, h_{t-1}) \quad (8)$$

   ii. Concise case level in-depth semantics as a whole in $h_t$ by applying GlobalMaxpooling to make it more informative and forms the second outcome of this module.

   $$Th = GlobalMaxpooling(h_t) \quad (9)$$

---

The first module named case level hierarchical encoder draws the task specific in-depth semantics for each case dynamically at multiple levels i.e, chunk and document level by following the divide and conquer approach as shown in Figure 2. The proposed algorithm for case level hierarchical encoder is shown in Table 1. The algorithm takes fact description as input and returns two outcomes named aggregated chunk and document semantics representing the first two stages in the proposed architecture as shown in Figure 2.

In the first stage named *Extracting chunk level semantics* the first component chunking takes the FD as input and iteratively segments it into the fixed sized chunks and then add CLS embedding for each chunk as delimiter. The resultant fixed sized chunks are then passed to the *Chunk Encoder* which might be a pretrained language model such as

XLNET transformer or encoder and return the embeddings for the input chunks. The resultant chunk embeddings are then passed to the *Chunk Semantics Extraction* component which is a combination of BiGRU and Maxpooling layers. The gate mechanism BiGRU defines the long-term dependencies between the chunk embedding vectors. The Maxpooling layer condense these chunk embedding vectors and returns concise individual chunk semantics like chunk summary which is the first outcome of the case level hierarchical encoder.

In the second stage named *Extracting document level semantics*, the individual chunk embedding vectors from first stage are concatenated to form aggregated chunk semantics It then passes it through the BiGRU gate to define long-term dependencies between these vectors. The GlobalMaxPooling layer condense these aggregated semantics as a whole and return the document level in-depth semantics which is the second outcome of this module.

*4.2 Concise extraction module:*

Generally, original court case document consists of the illustration of the case events and summary in the last few lines. Concise extraction module creates a concise view of the fact description using the extracted semantics from the first module. It concatenates the aggregated chunk semantics $h_k$ followed by the document semantics $Th$ as per the court case document structure having the document summary in the last few lines. Then applies attention on the concise document $Th^c$ to concentrate on the main words $Th^a$ impacting subsequent prediction module.

$$Th^c = concat(\ h_k\ ,Th) \qquad (10)$$

$$Th^a = attention(Th^c) \qquad (11)$$

*4.3 Prediction module:*

It uses the output from concise extraction module to make the final outcome prediction through sigmoid.

$$Y^p = sigmoid\ (W_t Th^a + U_t h_{t-1} + b_t) \qquad (12)$$

Here $W_t$ is the input weight; $U_t$ is the hidden weight; $h_{t-1}$ is the previous hidden state; $b_t$ is the bias. $Y^p$ is the task predicted distribution. Applied cross-entropy for loss function given by,

$$L = -\lambda \sum_{i=1}^{|Y|} Y_i log\ (Y_i^p) \qquad (13)$$

where $|Y|$ means the number of labels of the prediction task which is 100 for ILSI dataset and 2 for ILDC dataset, $Y_i$ is the ground-truth value of the label $i$, $Y_i^p$ is the predicted value of the label $i$, and $\lambda$ is the weight factor.

## V. EXPERIMENTAL RESULTS AND ANALYSIS

*5.1 Indian legal datasets for legal research*

The initial thought is to create a dataset with IPC sections to solve the given criminal cases. Regarding this visited multiple courts in the state of Andhra Pradesh, India and gathered data from multiple lawyers practicing in district courts located in the couple of cities. The main topics are the application of IPC sections, IRPC procedures, evidence act, etc., regarding the eight code books belong to criminal cases. Also discussed criminal case solving procedure, judge methodology in decision making, and the feasibility of creating dataset with IPC sections as initial approach to solve LJP multi task problem. But after the interview, found that IPC is just reference for sections and to know maximum and minimum penalty and prison terms possible. IRPC will give the case solving procedure and evidence act tells the evidences to consider and need to be proved for the judgment delivery. Also enlightened that IPC is the main code book containing sections common for all types of crimes. For detailed subsections corresponding code book in the remaining seven code books listed as IRPC, evidence act, the narcotic drugs and psychotropic substances (NDPS) act, the prevention of corruption act, the protection of children from sexual offences (POCSO) act, the Juvenile justice act, the unlawful activities (prevention) act need to be referred for respective acts as per the crime event happened e.g., drugs, smuggling. Although IPC and other code books gives the reference to possible prison and penalty term limits, the exact prison or penalty term decision to impose on the criminal is taken by the judge. Here he considers his case analysis, its severity and the evidences proved by following the IPC and other code books subsection punishment ranges. This is the methodology followed by the judge in the court for judgment decision making and case disposal.

So, we realized that making dataset with IPC sections alone will not give the exact solution to the case judgment decision. The traditional deep learning method of making the dataset and training the system with the previously solved cases is the feasible approach to simulate the judge decision and also adapted here to solve LJP. Also found that there are no standard datasets available for India to support various tasks of LJP [1]. So used available datasets of ILDC [5] and ILSI [4] to test our proposed SLJP model performance.

*5.2 Evaluation Metrics:*

Precision, recall, F1score and classification accuracy are the evaluation measures applicable to LJP. Definition of these metrics will differ based on the type of classification problem chosen. Here tested the SLJP model performance with two available Indian datasets named Indian Legal Statue Identification (ILSI) dataset [4] and Indian Legal Document Corpus (ILDC) dataset [5].

*Multi-label Classification:*

In multi-label classification, the classifier assigns multiple labels or classes simultaneously to a single input case document. Application of the proposed SLJP model on ILSI dataset is a multi-label classification problem which identifies applicable statues for the given case document. Here each case document is assigned with the multiple labels named applicable law sections as per the IPC based on the various crime events happened and proved in the case description.

Although calculated both label-based metrics of micro and macro average weights, here mainly considered the macro measures of precision (P), Recall (R), F1 score (F1), and classification binary Accuracy (A) to evaluate multi-label classification. The micro-averaging calculates each label as an individual prediction, whereas the macro-averaging calculates the average of per-label results as shown in the below equations 14, 15 for precision. In a similar way, recall will be calculated but divided by all the true cases ($TP_l+FN_l$). F1 score is calculated as the harmonic mean of the precision and recall [18] as shown in the equation 16. More precisely, the micro-average precision and recall score is calculated from the individual classes True Positives (TPs), True Negatives (TNs), False Positives (FPs), and False Negatives (FNs) of the model. The macro-average precision and recall score is calculated as the arithmetic mean of individual classes precision and recall scores. The macro-average F1-score is calculated as the arithmetic mean of individual classes F1-score[2]. Classification binary accuracy (A) is the arithmetic mean of the individual cases binary accuracy calculated by label wise binary mapping. The overall performance of a multi-label classifier mainly depends on the right selection of the threshold value applied on the predicted probabilities to convert those to binary values. SLJP mainly considered the macro averaging results for comparison.

$$Micro - P = \frac{\sum_{l=1}^{L} TP_l}{\sum_{l=1}^{L} TP_l + FP_l} \quad (14)$$

$$P = Macro - P = \frac{1}{|L|} \sum_{l=1}^{L} \frac{TP_l}{TP_l + FP_l} \quad (15)$$

$$Macro - F_1 = \frac{(2PR)}{(P+R)} \quad (16)$$

*Binary classification:*

In binary classification, the classifier assigns a label or class to an input case document out of the two possible outcomes. Application of the proposed SLJP model on ILDC dataset is a binary classification problem which predicts the case acceptance based on the various crime events given in the fact description. Given a case proceeding case document from the supreme court of India (SCI), the task of SLJP is to automatically predict the decision for the given case whether it is accepted or denied.

Although calculated both metrics of micro and macro average weights, here mainly considered the macro weights of precision (P), Recall (R), F1 score (F1), and classification Accuracy (A) to evaluate the ILDC dataset. Precision is the ratio of count of true positive results predicted to the count of all positive results predicted. Recall is the ratio of count of true positive results predicted to the actual count of positive samples to be predicted. The F1 score is the harmonic mean of the precision and recall and it act like a measure of test accuracy. Classification accuracy (A) is the ratio of number of cases predicted correctly to the total number of cases in the dataset. We mainly considered the macro averaging results for comparison.

*5.3 Proposed SLJP model results:*

Table 2 shows the proposed SLJP method performance on the various encoder and transformer embeddings. Here tested the performance for three encoding methods such as 1. *Sent2Vec* with input dimension of 300 for both datasets 2. *GloVe* with input dimension of 100 for both datasets 3. *Doc2Vec* with input dimension of 512 for ILSI dataset and 1000 for ILDC dataset. If the dataset is small, running the model multiple times will increase the performance. Among the three encoding methods Doc2Vec has shown the highest performance which is 60.77% accuracy on ILDC dataset for first run and F1 score of 49.47 during the second run on ILSI dataset for 20 epochs.

Also applied the proposed work on two transformer models named XLNet and BERT. *XLNET* takes input dimension of 512 features for ILSI dataset which had produced 568 chunks as maximum and 768 features for ILDC dataset and produced 25 chunks as maximum per case document. *BERT* takes the input dimension of 512 features for ILSI dataset and produced 516 chunks as maximum and 768 features for ILDC dataset and produced 25 chunks as maximum. Hence these transformers produced various number of chunks for the same input features on the datasets based on the maximum length of the record. Out of all SLJP with XLNet embeddings has shown the highest performance on both datasets which is 49.53% F1 score on ILSI dataset and 74.8% accuracy on ILDC dataset over various transformer and encoding methods for 20 epochs as shown in Figure 3 and 4 and Table 2. The proposed SLJP model has recorded the high performance with both transformers such as XLNet and BERT embeddings than the encoders as shown in the Table 2.

Assumed 512 features for ILSI dataset and 768 features for ILDC dataset for transformer models and various sizes for encoders based on the its available versions to draw embeddings. Tested all input embeddings for both data sets for 20epochs and applied standard scalar standardization technique for data preprocessing. Also tested Minmax scalar normalization data preprocessing technique for all the models on both datasets, but it has not given expected performance. Other hyper parameters applied are L2 regularization, Adam optimizer, 0.95 momentum, attention mechanism, global max pooling layer at document level semantics drawing, concise extraction module to enrich the semantics prior to the prediction, AUC-ROC curves as alternative technique for threshold determination, tested gradient changing technique too but it is increasing the running time and no much performance improvement, applied binary accuracy for multi label classification, applied shuffling of records during training to avoid overfitting, dropout 0.5, taken batch size 32 for ILSI dataset and 64 for ILDC dataset, ReLU activation function, used sigmoid function in dense layer.

Tested the SLJP performance with sigmoid in dense layer and tanh as activation function on ILDC dataset. It gives the slight performance improvement of 0.3% i.e., 74.8% accuracy which is the highest performance. And also tested the performance with SoftMax in dense layer and tanh as activation function on ILSI dataset, but it slightly degrades the performance to 48.96% F1 score as shown in the Table 3. Applied various learning rates to different layers in the model. For ILDC dataset applied 1e-4 for BiGRU, and 1e-5 for Maxpooling at chunk level, 1e-4 for BiGRU, and 1e-5 for GlobalMaxpooling at document level, and 3e-5 for attention layer. For ILSI dataset applied 1e-5 for BiGRU, and Maxpooling at chunk level, 1e-5 for BiGRU, and GlobalMaxpooling at document level, and 3e-5 for attention layer.

---
[2] https://stackoverflow.com/questions/68708610/micro-metrics-vs-macro-metrics

**Table 2:** SLJP results for various transformer and encoder methods ran for 20 epochs on two datasets. SLJP with XLNet embeddings has shown the highest performance on ILDC dataset with batch size 64. BERT and XLNET embeddings with SLJP given the highest and equal performance for ILSI dataset with batch size 32. Shown the results of macro average measure as matric for single run.

| S. No | Embedding Method | Model | ILSI Dataset (Batch size=32, Epochs=20) | | | | ILDC Dataset (Batch size=64, Epochs=20) | | | |
|---|---|---|---|---|---|---|---|---|---|---|
| | | | P | R | F1 | A | P | R | F1 | A |
| 1 | Transformers | BERT | 49.83 | 50.28 | **49.53** | 94.82 | 71.98 | 71.93 | 71.92 | 71.94 |
| 2 | | XLNet | 50.58 | 50.28 | **49.53** | 95.43 | 74.53 | 74.53 | 74.53 | **74.53** |
| 3 | Encoders | Doc2Vec | 49.43 | 50.16 | 49.47 | 95.09 | 60.8 | 60.77 | 60.74 | 60.77 |
| 4 | | GloVe | 49.36 | 50.49 | 49.46 | 94.92 | 25.03 | 50 | 33.36 | 50.06 |
| 5 | | Sent2Vec | 49.2 | 50.05 | 49.16 | 95.95 | 25.03 | 50 | 33.36 | 50.06 |

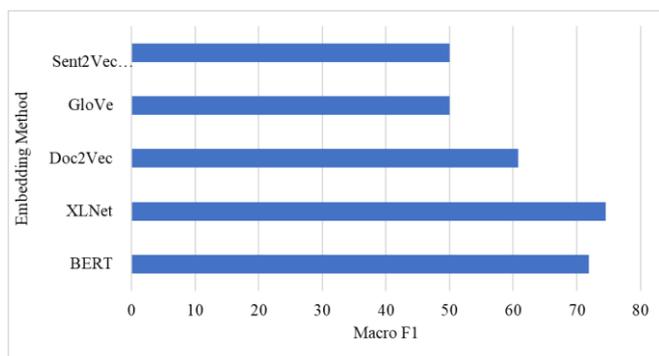

**Figure 3:** SLJP results with various embedding methods on ILDC dataset for 20 epochs.

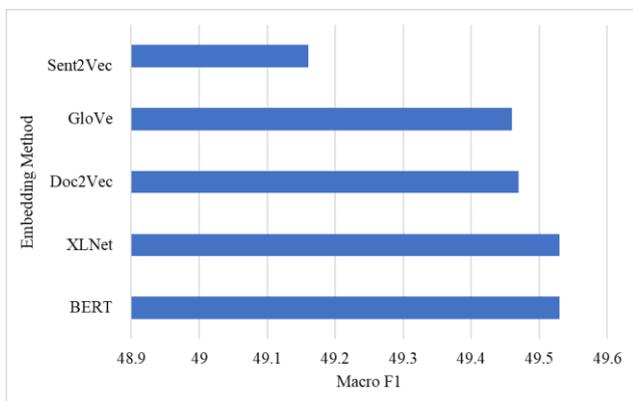

**Figure 4:** SLJP results with various embedding methods on ILSI dataset for 20 epochs.

SLJP results with various embedding methods on ILDC and ILSI datasets for 20 epochs has shown in Figure 3 and 4.

It is observed that, as the number of model layers increases more data loss occurs in processing consequently performance will degrade. The simple model with few basic layers such as LSTM, GRU gives the good performance. Table 4 discusses the comparison of SLJP model with other baselines shown the state-of-the-art performance so far. The proposed SLJP model shown the highest performance on both datasets than the baselines. Tested the SLJP performance with Indian pretrained transformers, but not given the expected results.

Considered the following embedding methods and libraries to transform the raw case document text into numerics: adapted the FastText wikipedia unigram model as Sent2Vec [6] technique to produce embeddings for both datasets[3], used inbuilt python library named gensim to produce Doc2Vec [7] embeddings implementing paragraph vector logic for both datasets[4], applied gensim's library glove-100d to produce GloVe [8] embeddings for both datasets[5]. For BERT and XLNet transformer embeddings adapted the methods discussed in the works [9, 10] for both datasets.

Also implemented the two types of concise extraction module logics: 1) First is concatenating aggregated chunk and document semantics along with attention mechanism on the top 2) Second is concatenating the document semantics and raw input embeddings along with attention mechanism on the top. But the first logic has shown the good performance than unprocessed raw input embeddings in the second logic adapted the first logic as CE module in the model. Tested the experimental results using NVIDIA GeForce GTX 1080 24GB GPU system. Table 4 compares the SLJP model with baseline models shown the state-of-the-art performance on both datasets. Noted the SLJP model with XLNet embeddings exhibited the highest performance on both datasets compared to the baselines.

### 5.4 Performance analysis of SLJP:

This section analysis the performance of the proposed SLJP model on both available Indian datasets and its baselines.

*ILSI Dataset:*

Application of the proposed SLJP model on ILSI dataset is a multi-label classification problem. Here each case document is assigned with the multiple law sections as per the IPC based on the various crime events happened in the case description. ILSI dataset [4] consists of cases referring top 100 frequently occurred sections from IPC. Table 5 shows the performance of SLJP model comparing with other baselines on ILSI dataset. Considered the works [4,11,12,13,14,15,16,17] discussed in the related work section as baselines. Figure 5 shows the comparison of SLJP performance with baseline models of ILSI dataset graphically. The proposed SLJP model has recorded the highest performance with transformers such as XLNet and BERT embeddings than its baselines as shown in the Figure 5. Results of the baseline models are taken from the work [4] for comparison.

**Table 3:** SLJP performance with highly performed XLNet embeddings with various hyper parameters for single run, 20 epochs. Recorded the highest performance on ILDC dataset than normal hyper parameters applied commonly to various embedding methods in table 2.

| S. No | Embedding Method | Dataset | Hyper parameters | P | R | F1 | A |
|---|---|---|---|---|---|---|---|
| 1 | XLNet | ILDC | Tanh activation, Sigmoid dense layer | 74.89 | 74.8 | **74.77** | **74.8** |
| 2 | | ILSI | Tanh activation, SoftMax dense layer | 48.04 | 0.5 | 48.96 | 96.09 |

**Table 4:** Baseline models shown the state-of-the-art performance on both datasets. Shown the results of macro average measure. SLJP model with XLNet embeddings shown the highest performance on both datasets compared to the baselines.

| S. No | Dataset | Model | Epochs | P | R | F1 | A |
|---|---|---|---|---|---|---|---|
| 1 | ILDC | **XLNet+BiGRU [1]** | 20 | 68.8 | 68.8 | 68.8 | 68.79 |
| 2 | | **SLJP** | 20 | 74.53 | 74.5 | **74.53** | 74.53 |
| 3 | | **LeSICiN [2]** | 100 | 27.9 | 31.3 | 28.45 | - |
| 4 | ILSI | **SLJP** | 20 | 50.58 | 50.3 | **49.53** | 95.43 |

Implemented the dynamic threshold determination technique to learn a robust threshold for each label automatically. As ILSI is a multi-label dataset, determined the label wise thresholds that gives best F1 score for that label to convert model predictions to binary values. Performance metrics such as precision, recall, F1 score, and accuracy will be evaluated based on these binary values. Here performance comparison is done based on the calculated F1 score and considered macro measures.

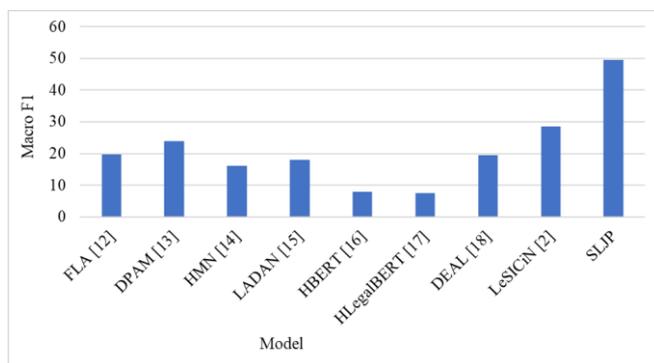

**Figure 5:** Comparison of SLJP performance with baseline models of ILSI dataset.

*ILDC Dataset:*

Application of the proposed SLJP model on ILDC dataset is a binary classification problem which predicts the case acceptance based on the various crime events given in the fact description. ILDC dataset [5] consists of the supreme court of India (SCI) case proceeding documents where the task of SLJP is to dynamically predict the decision for the given case whether it is accepted or denied. Table 6 shows the performance of SLJP model comparing with other base models tested on ILDC [5] dataset for various number of epochs. Out of all the base models XLNet+BiGRU has shown the highest performance and adapted it to compare with SLJP results.

**Table 5:** Comparison of SLJP performance with baseline models of ILSI dataset. SLJP has recorded the highest performance with XLNet and BERT embeddings than the baselines.

| S.No | Model | Epochs | P | R | F1 |
|---|---|---|---|---|---|
| 1 | FLA [11] | 100 | 12.19 | 69.25 | 19.6 |
| 2 | DPAM [12] | 100 | 27.11 | 27.43 | 23.86 |
| 3 | HMN [13] | 100 | 10.27 | 57.3 | 16.12 |
| 4 | LADAN [14] | 100 | 12.54 | 46.17 | 17.93 |
| 5 | HBERT [15] | 100 | 5.79 | 51.43 | 7.91 |
| 6 | HLegalBERT [16] | 100 | 4.36 | 52.55 | 7.58 |
| 7 | DEAL [17] | 100 | 12.66 | 64.83 | 19.43 |
| 8 | LeSICiN [4] | 100 | 27.9 | 31.32 | 28.45 |
| 9 | **SLJP** | **20** | **50.58** | **50.28** | **49.53** |

The proposed SLJP model has recorded the highest performance with XLNet transformer embeddings than base models for various number of epochs e.g., 2,3,4,20. Results of some baseline models are taken from the work [5] for comparison and left over is the tested outcomes. Also noted the less performance in our environment setup than given in the work [5] for the same base model codes made available. So, ran the base model codes for various epochs in our environment for comparison. For example, XLNet+BiGRU and XLNet+BiGRU+Attention model ran for 2 epochs showed the varied tested results in the Table 6.

---
[3] https://github.com/celento/sent2vec
[4] https://radimrehurek.com/gensim/models/doc2vec.html
[5] https://zenodo.org/records/4925376

**Table 6:** Comparison of SLJP model with the base models tested on the ILDC dataset for various epochs.

| S. No | Methodology | | epochs | P | R | F1 | Accuracy |
|---|---|---|---|---|---|---|---|
| | Input Embeddings | Model | | | | | |
| 1 | Sent to Vec | BiGRU+Att. [5] | 1 | 60.98 | 58.4 | 59.66 | 58.31 |
| | | SLJP | 20 | 25.03 | 50 | 33.36 | 50.06 |
| 2 | Doc to Vec | BiGRU+Att. [5] | 2 | 57.18 | 56.03 | 56.6 | 57.44 |
| | | SLJP | 20 | 60.8 | 60.77 | 60.74 | 60.77 |
| 3 | Glove | BiGRU+Att. [5] | 3 | 68.26 | 60.87 | 64.35 | 60.75 |
| | | SLJP | 20 | 25.03 | 50 | 33.36 | 50.06 |
| 4 | Bert | Base [5] | 3 | 69.33 | 67.31 | 68.31 | 67.24 |
| | | BiGRU [5] | 3 | 70.98 | 70.42 | 70.69 | 70.38 |
| | | BiGRU+Att. [5] | 2 | 71.31 | 70.98 | 71.14 | 71.26 |
| | | SLJP | 20 | 71.98 | 71.93 | 71.92 | 71.94 |
| 5 | XLNET | Base [5] | 3 | 72.09 | 70.07 | 71.07 | 70.01 |
| | | BiGRU | 2 [5] | 77.8 | 77.78 | 77.79 | 77.78 |
| | | | **2** | **75.3** | **74.3** | **74.66** | **74.07** |
| | | | 3 | 74.96 | 73.03 | 73.98 | 73.08 |
| | | | 4 | 74.68 | 73.57 | 74.12 | 73.61 |
| | | | **20** | **68.8** | **68.8** | **68.8** | **68.79** |
| | | BiGRU+Att. | 2 [5] | 77.32 | 76.82 | 77.07 | 77.01 |
| | | | 2 | 75.14 | 73.49 | 74.31 | 73.54 |
| | | SLJP | **2** | **75.47** | **75.46** | **75.46** | **75.46** |
| | | | 3 | 76.25 | 76.13 | 76.19 | 76.13 |
| | | | 4 | 76.18 | 75.85 | 76.01 | 75.86 |
| | | | **20** | **74.53** | **74.53** | **74.53** | **74.53** |

**Table 7:** Performance variation of SLJP and BiGRU model for increased epochs with highly performed XLNet embeddings.

| S. No | Embedding Method | Epochs | BiGRU Model | SLJP Model |
|---|---|---|---|---|
| 1 | XLNet | 2 | 74.66 | 75.46 |
| | | 3 | 73.98 | 76.19 |
| | | 4 | 74.12 | 76.01 |
| | | **20** | **68.8** | **74.53** |

Observed the decreasing performance of base models in [5] with the increase of epoch count. In SLJP model the decreasing rate is very less comped to the base models in [5]. Portion of the Table 6 is shown in the Table 7 to show the performance variation of SLJP and BiGRU model for increased epochs with highly performed XLNet embeddings. Its graphical representation is shown in Figure 6. Noted the less performance degradation in the proposed SLJP model than BiGRU model for increased epochs.

In ILDC paper [5] considered last 512 words of the document as FD which will form single chunk. Even though architecture of XLNet+BiGRU shows the iteration over multiple chunks, in real case iteration is not implemented because of last 512 words consideration. Whereas SLJP model considered the entire document as FD rather than last 512 words and implemented chunk wise iteration to draw chunk wise semantics. Also proved that considering the entire document as FD will improve the performance and performance decreasing rate is also less as number of epochs increases compared to the base models in [5].

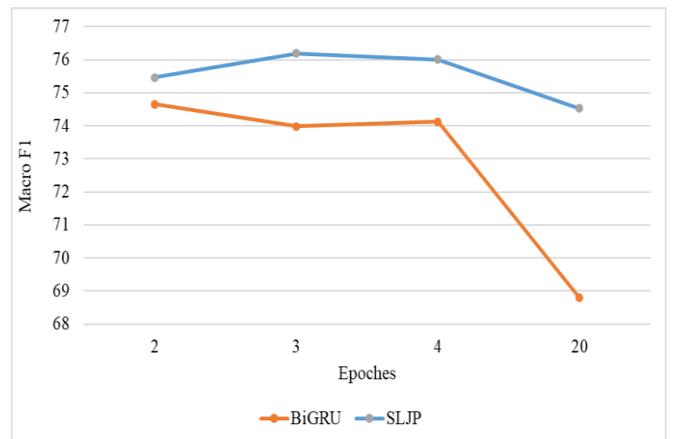

**Figure 6:** Graphical representation of performance variation of the SLJP and BiGRU model for increased epochs with highly performed XLNet embeddings. Noted the less performance degradation in SLJP than BiGRU model as epochs increases.

Implemented the *dynamic threshold determination technique* to learn a robust threshold automatically. As ILDC is a binary dataset, determined the threshold value that gives the best F1 score on predictions which converts model predictions to binary values. Performance metrics precision, recall, F1 score, and accuracy will be evaluated based on these binary values. Here performance comparison is done based on the calculated F1 score and accuracy.

## VI. ABLATION ANALYSIS

### 6.1 Transformer selection:

Analyzed the performance of two popular transformers named BERT and XLNET with a combination of BiGRU + attention technique for LJP prediction on the ILDC-multi dataset as shown in the Table 8. XLNET performed better than BERT, adapted XLNET pretrained language model for SLJP model design. Considered the macro weights for results evaluation.

**Table 8:** Comparision of BERT and XLNet transformers performance on ILDC dataset [5].

| S. No | Model | P | R | F1 | A |
|---|---|---|---|---|---|
| 1 | BERT+BiGRU+att. | 71.31 | 70.98 | 71.14 | 71.26 |
| 2 | XLNet+BiGRU+att. | 77.32 | 76.82 | 77.07 | 77.01 |

### 6.2 Gate selection:

After selecting the XLNET the performance of it is analyzed on various gate mechanisms such as LSTM, GRU, BiLSTM, and BiGRU along with the attention mechanism on ILDC dataset for judgment prediction. The BiGRU exhibits the highest performance as shown in the Table 9 and likely to adapt in the proposed SLJP architecture to learn the dependencies between the data in chunk and document semantics extraction.

**Table 9:** Performance of XLNET with various gate mechanisms on ILDC dataset.

| S. No | Model | P | R | F1 |
|---|---|---|---|---|
| 1 | XLNet + BiGRU + att. | 77.32 | 76.82 | **77.07** |
| 2 | XLNet + GRU + att. | 74.55 | 74.32 | 74.43 |
| 3 | XLNet+BiLSTM+ att. | 75.98 | 75.91 | 75.94 |
| 4 | XLNet + LSTM + att. | 74.95 | 74.37 | 74.66 |

### 6.3 Importance of various modules in SLJP model:

Table 10 shows ablation analysis of SLJP model on ILSI and ILDC datasets. Analyzed the importance of each critical component involved in the SLJP architecture such as chunk semantics, document semantics and concise extraction module by testing the model performance in its absence on both datasets. From the Table 10 observed the performance of SLJP method degrades after eliminating either component for ILSI dataset. For ILDC dataset without document semantics shown the highest performance because ILDC dataset contains last few sentences identified as document summary from case document as fact description. So, summarizing it further at document level reduces the performance slightly because of more loss of semantics. Through this experiment observed that increasing the count of processing steps on raw data through number of model layers will decrease the performance of the model.

**Table 10:** Ablation analysis of SLJP model on ILSI and ILDC datasets to analyze the importance of various components in SLJP model. Below table shown the results of test accuracy belongs to both datasets.

| S.No | XLNet Model Embeddings | ILSI | ILDC |
|---|---|---|---|
| 1 | With out Chunk level semantics | 84.1 | 74.53 |
| 2 | With out document level semantics | 82.65 | **75.66** |
| 3 | With out Concise Extraction module | 84.27 | 71.8 |
| 4 | SLJP | **87.7** | 74.53 |

## VII. CONCLUSION AND FUTURE WORK

The proposed SLJP model provides the advantages of using pretrained language models named transformers for complex unstructured legal case document understanding and to generate embeddings. It draws the in-depth semantics of the given fact description at multiple levels i.e., chunk and case document level by following the divide and conquer approach. It creates concise view of the fact description as per the original court case document structure having document summary at the last few lines and predicts judgment using attention mechanism. Successfully tested the SLJP performance on both binary and multi-label classification to predict the judgment.

Found the lack of multi-tasking dataset for Indian context to predict three legal subtasks e.g., law articles, prison term, and penalty term. So, SLJP was limited to available ILSI dataset for identifying applicable law statues. The proposed SLJP model applies hierarchical architecture in defining the dependencies between the fact description and statues and got promising results over ILSI dataset. In future SLJP will be extended to predict the other two judgment components such as prison and penalty term. Also applied the proposed SLJP model on other available binary classification dataset of ILDC and shown the highest performance than XLNET+BiGRU base model, previously shown highest performance on ILDC [5] for 20 epochs. It is proved that performance degradation in SLJP is less compared to the tested base models in [5] on ILDC as number of epochs increases. Also proved that considering the entire document rather than last few lines as FD will improve the performance.

ACKNOWLEDGEMENTS

The authors would like to thank Mr. Mallikarjun Rao, advocate, practicing in the Tirupati district court and advocates in Srikalahasti district court. They enlightened us on the application of IPC, methodology followed by the judge in verdict delivery, and the feasibility of making the dataset with IPC to solve criminal cases. The information helped us in taking the critical decision of dataset selection and to move forward in evaluating the SLJP approach. This discussion realized us on testing the SLJP model with datasets consisting of previously solved cases named precedents which is the right approach than creating the dataset with IPC sections.